\pdfoutput=1
\documentclass{article}

\usepackage[utf8]{inputenc}
\usepackage[T1]{fontenc}
\PassOptionsToPackage{hyphens}{url}
\usepackage[breaklinks=true,hidelinks]{hyperref}
\usepackage{url}
\usepackage{booktabs}
\usepackage{amsfonts}
\usepackage{amsmath}
\usepackage{amssymb}
\usepackage{nicefrac}
\usepackage{microtype}
\usepackage{graphicx}
\usepackage[dvipsnames]{xcolor}
\usepackage{subcaption}
\usepackage{natbib}
\usepackage{listings}
\usepackage{float}
\usepackage{longtable}
\usepackage[ruled,vlined]{algorithm2e}
\usepackage[most]{tcolorbox}

\SetCommentSty{mycommfont}
\SetKwComment{Comment}{// }{}
\setcitestyle{authoryear,round,citesep={;},aysep={,},yysep={;}}

\definecolor{promptblue}{RGB}{41,98,150}
\definecolor{promptbg}{RGB}{245,248,252}

\definecolor{codebg}{HTML}{F7F7F5}
\definecolor{codeframe}{HTML}{E2E5E9}
\definecolor{codetext}{HTML}{2F343D}
\definecolor{codenum}{HTML}{A7ADB5}

\definecolor{codekeyword}{HTML}{6E6CCF}  
\definecolor{codecomment}{HTML}{A6ADB8}  
\definecolor{codestring}{HTML}{B38AD8}   
\definecolor{codefunc}{HTML}{B46A7A}     
\definecolor{codeconst}{HTML}{5E9C93}    

\lstdefinestyle{acm_style}{
    backgroundcolor=\color{codebg},
    basicstyle=\ttfamily\footnotesize\color{codetext},
    commentstyle=\color{codecomment},
    keywordstyle=\bfseries\color{codekeyword},
    stringstyle=\color{codestring},
    identifierstyle=\color{codetext},
    numberstyle=\scriptsize\color{codenum},
    numbers=left,
    numbersep=8pt,
    xleftmargin=4pt,
    frame=single,
    rulecolor=\color{codeframe},
    framerule=0.4pt,
    framesep=5pt,
    breaklines=true,
    breakatwhitespace=false,
    showspaces=false,
    showstringspaces=false,
    showtabs=false,
    keepspaces=true,
    tabsize=4,
    captionpos=b,
    aboveskip=6pt,
    belowskip=6pt,
    emph={range,len,min,max,sum,zip,enumerate,print,abs,sorted,np,array,zeros,ones,clip,tile,uniform},
    emphstyle=\color{codefunc},
    emph={[2]True,False,None},
    emphstyle={[2]\color{codeconst}}
}

\newtcolorbox{promptbox}[1][]{
  enhanced,
  title=#1,
  fonttitle=\bfseries\small,
  colback=promptbg,
  colframe=promptblue!70,
  colbacktitle=promptblue!85,
  coltitle=white,
  arc=3pt, boxrule=0.5pt,
  top=4pt, bottom=4pt, left=6pt, right=6pt,
  fontupper=\small,
}

\title{Evolving Ensemble of Agents}

\author{
  Zongmin Yu \\
  National University of Singapore \\
  \texttt{yuzongmin@u.nus.edu} \\
  \and
  Liu Yang\thanks{Corresponding author.} \\
  National University of Singapore \\
  \texttt{yangliu@nus.edu.sg} \\
}

\date{}

\begin{document}

\maketitle

\begin{abstract}
We introduce the Evolving Ensemble of Agents (EvE), a decentralized framework that organizes existing, highly capable coding agents into a live, co-evolving system for algorithmic discovery. Rather than reinventing the wheel within the ``LLMs as optimizers'' paradigm, EvE fixes the base agent substrate and focuses entirely on evolving the cumulative guidance and skills that dictate agent behaviors. By maintaining two co-evolving populations, namely functional code solvers and agent guidance states, the system evaluates agents through a synchronous race, updating their empirical Elo ratings based on the marginal gains they contribute to the current solver state. When applied to a research bottleneck in In-Context Operator Networks (ICON), EvE autonomously discovered a robust rescale-then-interpolate mechanism that enables reliable example-count generalization. Crucially, controlled ablations reveal the absolute necessity of stage-dependent agent adaptation to navigate the shifting search landscapes of complex codebases. Compared to variants driven by a fixed initial agent or even a frozen ``best-evolved'' agent, EvE uniquely avoids phase mismatch, demonstrating that organizing agents into a self-revising ensemble is the fundamental driver for breaking through static performance ceilings.

\end{abstract}

\section{Introduction} \label{sec:intro}

The pursuit of automated scientific research and algorithmic discovery driven
by Large Language Models (LLMs) represents a profound ambition in artificial
intelligence. In recent years, LLM-based code-evolution systems, such as
FunSearch~\citep{romeraparedes2024funsearch},
AlphaEvolve~\citep{novikov2025alphaevolve},
OpenEvolve~\citep{algorithmicsuperintelligence2026openevolve},
CodeEvolve~\citep{codeevolve}, and
ShinkaEvolve~\citep{shinkaevolve2025}, have made exciting strides by showing
that program search can discover useful algorithms. However, despite these
achievements, their operational capabilities on complex codebases still fall
significantly short of those of modern coding agents.

\begin{figure}[t]
  \centering
  \includegraphics[width=\textwidth]{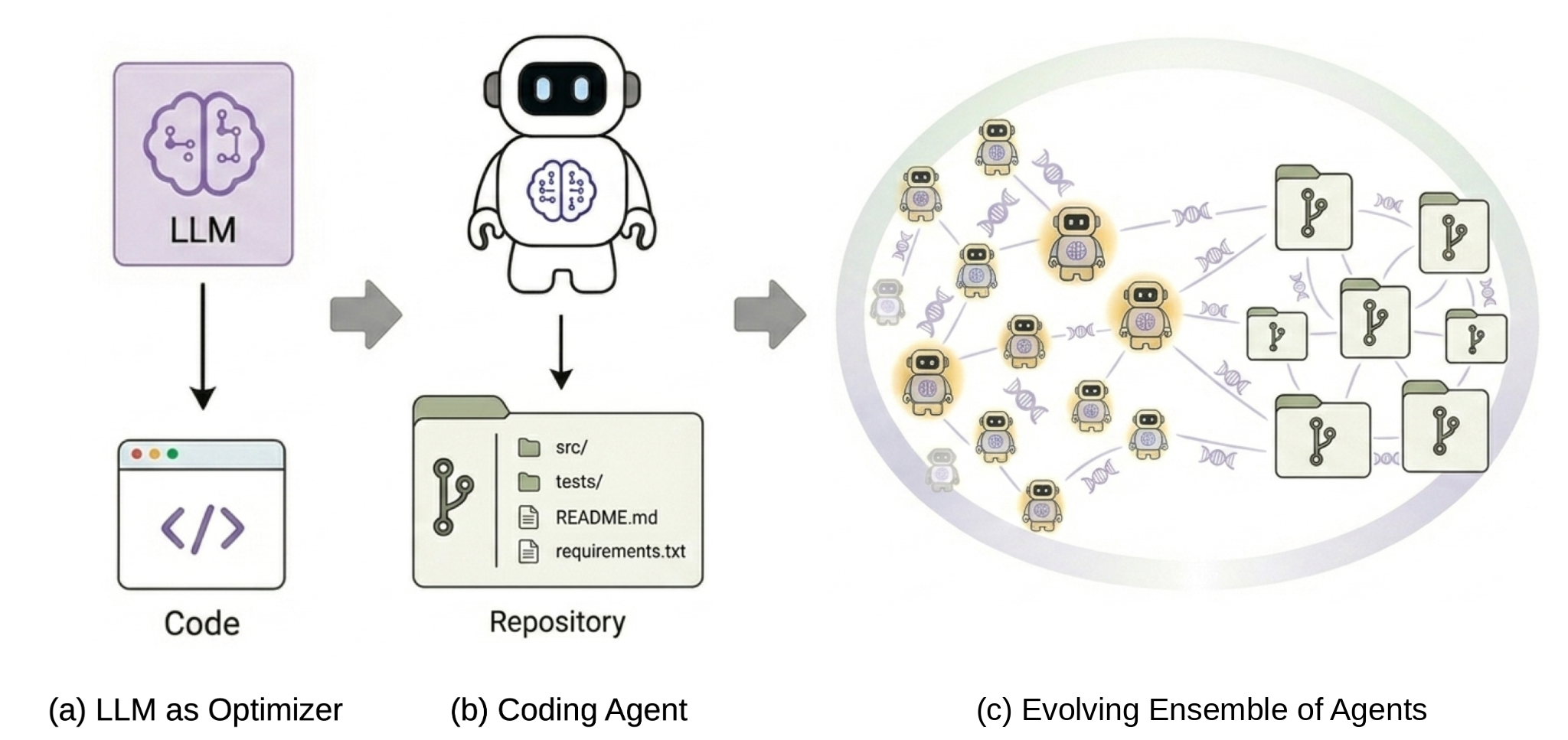}
  \caption{Three paradigms of LLM-driven algorithmic discovery.
    \textbf{(a) LLM as Optimizer.} An LLM proposes a code block, which is scored and re-prompted to LLM.
    \textbf{(b) Coding Agent.} A modern coding agent operates on a
    full code repository with autonomous planning, tool use, and
    sub-agent invocation.
    \textbf{(c) Evolving Ensemble of Agents (this work).} A decentralized
    ensemble of coding agents that evolves with another population of functional components within a code repository.}
  \label{fig:paradigms}
\end{figure}

Today, the capabilities of modern coding agents are already exceptionally
advanced. They can operate with autonomous planning, complex reasoning,
sophisticated context management, and sub-agent invocation, often without
human intervention. Because these foundational agentic capabilities are
already highly mature, we choose to shift toward organizing existing, highly capable coding agents, rather than reinventing the wheels.
In particular, we believe the challenge lies in scalable orchestration of
agents, knowledge sharing, and management of agent guidance and skills. 
In this paper, we present the Evolving Ensemble of Agents
(EvE) as an exploration in this direction
(Figure~\ref{fig:paradigms}).

We aim to develop an agent ensemble that continuously refines both downstream task solvers, represented as functional components within a code repository, and the agents themselves. Given that modern coding agents are already exceptionally capable, and their behaviors are predominantly steered by customized guidance and skills, we fix the base agent substrate and conceptualize the evolution of the guidance and skills as the evolution of the agents themselves. Notably, the EvE framework remains fully compatible should the base agent substrate be explicitly incorporated into the evolutionary loop.

In our design, EvE addresses the aforementioned challenges for organizing agents in an elegant way.

First, EvE unlocks scalable orchestration through a completely decentralized ensemble architecture. By abandoning rigid agent roles (e.g., ``leaders'' versus ``workers''), EvE achieves universal compatibility. In principle, any existing agent or multi-agent system can be seamlessly encapsulated as a single individual with the ensemble. More profoundly, this naturally supports recursive nesting, allowing an entire ensemble to function as an individual inside a higher-level ensemble. Consequently, EvE can adapt to various state-of-the-art systems with minimal structural modifications.

Second, knowledge sharing is seamlessly achieved across the agent ensemble
through an evolutionary substrate. This occurs on two interconnected levels.
On the first level, agents observe prior solver attempts generated by their
peers, allowing them to learn from both successes and failures. On the second
level, EvE continuously generates new guidance and skills, which
directly take existing agents and their working logs as references.

Finally, EvE realizes the autonomous management and evolution of agent
skills in a rigorous evolutionary way. Agents are encouraged to improve
guidance and skills while editing code repositories. This guidance and these
skills are then repeatedly evaluated during future coding edits, with concrete scores that drive sampling probability.

We applied EvE on a task encountered during our own research, in particular, 
the positional-encoding design for in-context operator networks (ICON)~\citep{yang2023icon,yang2024icon_pde,yang2025iconlm,cao2024vicon,wu2026gicon,zhang2025genicon,meng2025iconfinance,cole2026iconmeasure,liu2023iconrobust,cole2024icontheory,mishra2025continuum}. In an ablation study with controlled token budgets, EvE achieves the strongest, most stable performance compared to a fixed initial agent or ``best-evolved'' agent. These results validate that
organizing agents into a live, evolving ensemble is the fundamental driver of breaking through performance ceilings.
The repository for EvE and the generated artifacts for ICON are hosted at \url{https://github.com/scaling-group/eve}.

\section{Method}
\label{sec:algorithm}

\begin{figure}[t]
  \centering
  \includegraphics[width=\textwidth]{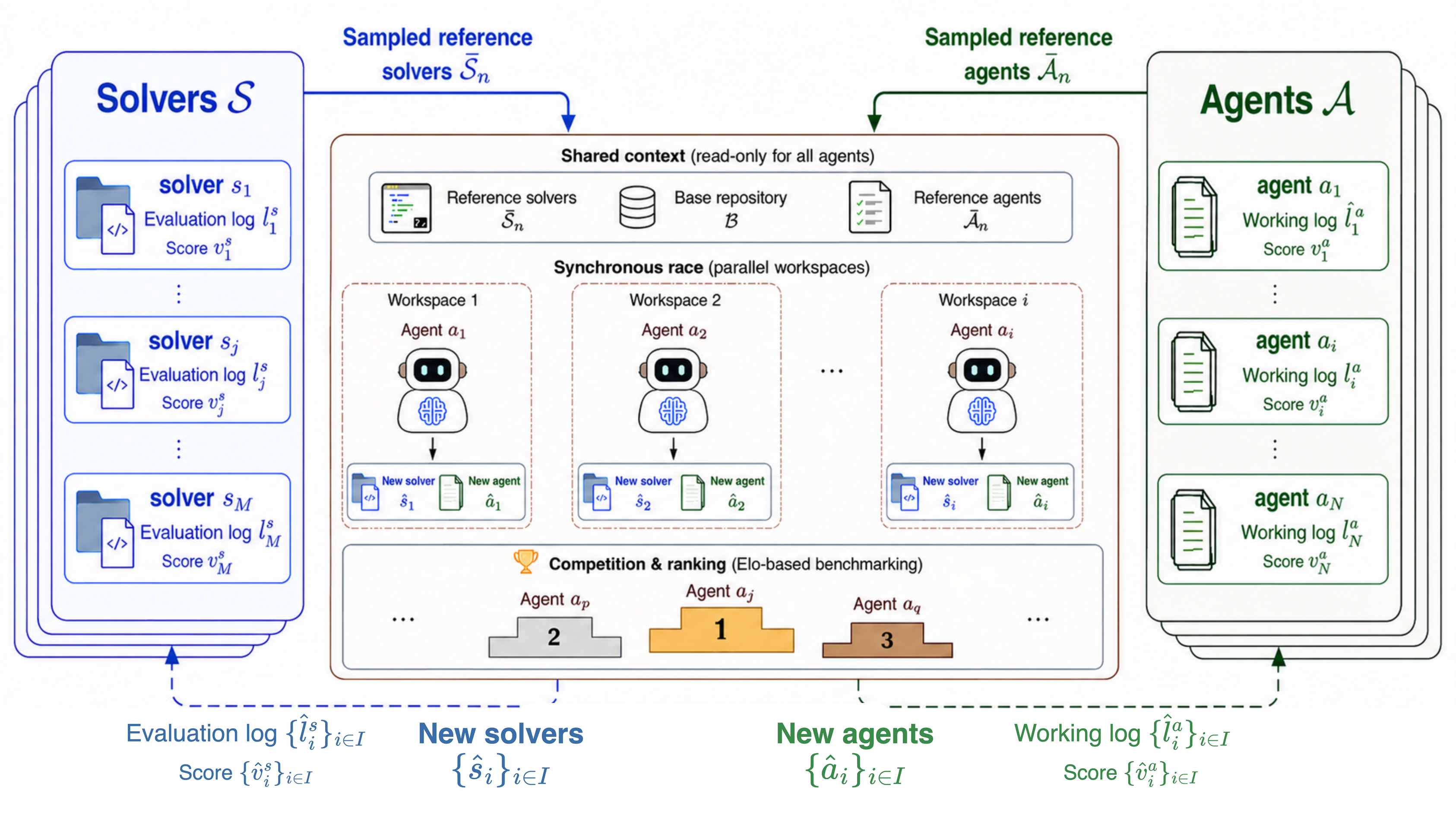}
  \caption{
  The EvE framework. EvE maintains two co-evolving populations: a solver population $\mathcal{S}$ containing functional components in a code repository, and an agent population $\mathcal{A}$ where each agent carries cumulative working logs and an Elo-based score. In each iteration, a ``synchronous race'' is conducted: multiple working agents $a_i$ are sampled to operate within mutually independent yet identical workspaces. While the environmental context, comprising reference solvers $\bar{\mathcal{S}}_n$, reference agents $\bar{\mathcal{A}}_n$, and the base repository $B$, is held constant, each agent utilizes its unique guidance and skills to produce a new solver $\hat{s}_i$ and a revised agent $\hat{a}_i$. This controlled setup allows agent Elo ratings to be updated based on the performance of their newly generated solvers, effectively distinguishing the effectiveness of each agent's strategy.}
  \label{fig:framework}
\end{figure}

\subsection{Agents and Solvers}

EvE maintains two scored populations (Figure~\ref{fig:framework}):
\[
    \mathcal{S}=\{(s_j, l^s_j, v^s_j)\}, \qquad
    \mathcal{A}=\{(a_i, l^a_i, v^a_i)\}.
\]
Here $s_j$ represents a solver, defined as a set of code files within the base repository $B$, associated with its evaluation log $l^s_j$ and score $v^s_j$ provided by the task evaluator $f$. $a_i$ denotes an agent with its cumulative working logs $l^a_i$ and a score $v^a_i$.

Scoring agents plays a critical role in the evolution of agents. It acts not only as a gatekeeper for sampling probability but also as a compass for agent improvement. The primary difficulty lies in the shifting requirements of the solver; much like a human organization, the ideal structure for a startup phase differs fundamentally from that of a mature enterprise. In alignment with the No Free Lunch theorem, which posits that no single optimization strategy excels across all problem domains, a universal agent is inherently sub-optimal across the entire lifecycle of algorithmic discovery. As the solver evolves from a ``startup'' phase to a ``late-stage refinement'' phase, the underlying search landscape shifts, requiring a corresponding evolution in agent guidance and skills. 

EvE manages this complexity by scoring the agents via a dynamic benchmarking mechanism, where the value of an agent is tied to its marginal contribution to the current states of solvers. Specifically, EvE ranks agents by treating each optimization iteration as a synchronous race. Multiple agents are sampled to refine the same set of high-performing solvers, and their relative Elo ratings are updated based on the quality of their respective outputs. This pairwise competition among agents focuses on their ability to extract marginal gains from an identical baseline, allowing the ensemble to identify and promote the most effective optimization strategies for the current developmental stage. This Elo-based scoring method is adopted from Escher-Loop \citep{escher_loop_2026}.

Specifically, in each iteration $n$, EvE samples a set of high-performing working agents $\mathcal{A}_n$, along with reference sets of solvers $\mathcal{\bar{S}}_n$ and agents $\mathcal{\bar{A}}_n$, which are combined with the base code repository $B$ to provide context for the evolutionary step. Each working agent $a_i \in \mathcal{A}_n$ operates on the same reference set to generate new solvers and agents. This execution is denoted as:
\begin{equation}
    (\hat{s}_i, \hat{a}_i, \hat{l}^a_i) = a_i(\mathcal{\bar{S}}_n, \mathcal{\bar{A}}_n, B).
    \label{eq:eve_worker_call}
\end{equation}
The outputs consist of a new solver candidate $\hat{s}_i$, a potentially revised agent guidance $\hat{a}_i$, and a session log $\hat{l}^a_i$ documenting the generation process. 

By forcing all sampled agents $a_i \in \mathcal{A}_n$ to refine the same reference set, EvE constructs a strictly pairwise competition where the variance in the resulting solver quality $\{\hat{v}^s_i\}_{i \in I}$ is directly attributed to the effectiveness of each agent's specific strategy. As detailed in Algorithm~\ref{alg:eve-search}, this process begins with the construction of a win-loss matrix $W$ based on the evaluation results of the solver candidates. These outcomes are then processed via the \text{EloUpdate} function to adjust the agents' scores $\{v^a_i\}_{i \in I}$. This approach isolates the marginal contribution of each agent relative to an identical baseline, allowing the ensemble to identify and promote the most effective optimization strategies for the current developmental stage.

Following evaluation, the agent ensemble is expanded by integrating modified agents $\hat{a}_i \neq a_i$ along with their session logs $\hat{l}^a_i$ into $\mathcal{A}$, ensuring that new optimization strategies and their underlying procedural evidence, including reasoning traces and failed attempts, are preserved.

\subsection{Integrated Agent Workspace}
\label{subsec:closed_loop_search}

Modern coding agents are extremely proficient at navigating complex file systems and manipulating source code within a structured environment. EvE leverages this by constructing a dedicated workspace for each agent, with all dependencies included in the workspace. The agent then performs a series of file-system operations to produce the refined solvers and agents, with its modification scope explicitly restricted to designated files and subsequently enforced by rigorous post-generation checks. This marks a departure from the early experimentation of Escher-Loop, which remains limited to the ``LLMs as optimizers'' paradigm by evolving code blocks and LLM prompt builders, with limited capability on complex tasks.

Instead of the rigid, alternating multi-phase process of the original Escher-Loop, EvE integrates solver improvement and self-referential agent optimization into a single, unified stage. In this way, EvE maximizes the scalability and parallelism of the entire evolutionary system. Furthermore, this integrated design provides a much broader context for both solver and agent refinement, compared with the decoupled multi-phase evolution in Escher-Loop. EvE grants the working agent full visibility into the examples of solvers and agents, their scores and logs, and the base code repository all at once, ensuring that the self-referential agent optimization undergoes more purposeful, context-aware evolution, directly adapting their strategies to the specific structure and optimization bottlenecks of the codebase they are improving.

\begin{algorithm}[t]
\caption{Evolving Ensemble of Agents}
\label{alg:eve-search}
\DontPrintSemicolon
\SetAlgoLined
\KwIn{Base code repository $B$, Task evaluator $f$; initial solver population $\mathcal{S}_0$; initial agent population $\mathcal{A}_0$; total iterations $T$.}
\KwOut{Updated populations $\mathcal{S}$ and $\mathcal{A}$.}
\BlankLine
$\mathcal{S} \gets \mathcal{S}_0$; $\mathcal{A} \gets \mathcal{A}_0$ \;
\For{$n = 1,\ldots, T$}{
    $\mathcal{A}_n = \{(a_i, l^a_i, v^a_i)\}_{i \in I} \gets \text{Sample}(\mathcal{A})$ \Comment*[r]{Sample working agents}
    $\mathcal{\bar{S}}_n =\{(s_j, l^s_j, v^s_j)\}_{j \in J} \gets \text{Sample}(\mathcal{S})$ \Comment*[r]{Sample reference solvers}
    $\mathcal{\bar{A}}_n = \{(a_k, l^a_k, v^a_k)\}_{k \in K} \gets \text{Sample}(\mathcal{A})$ \Comment*[r]{Sample reference agents}
    \BlankLine
    \For{$i \in I$ \text{in parallel}}{$(\hat{s}_i,\hat{a}_i,\hat{l}^a_i) \gets a_i(\mathcal{\bar{S}}_n, \mathcal{\bar{A}}_n, B)$ \Comment*[r]{Generate new solver and agent}
        $(\hat{l}^s_i, \hat{v}^s_i) \gets f(\hat{s}_i)$ \Comment*[r]{Evaluate solver}
        $\mathcal{S} \gets \mathcal{S} \cup \{(\hat{s}_i,\hat{l}^s_i, \hat{v}^s_i)\}$ \Comment*[r]{Add to solver population}
        $l^a_i \gets l^a_i \cup \hat{l}^a_i$ \Comment*[r]{Update agent working logs}
    }
    \BlankLine
    $W \gets \text{Competition}(\{\hat{v}^s_i\}_{i \in I})$ \Comment*[r]{Construct pairwise win-loss matrix}
    $\{v^a_i\}_{i \in I} \gets \text{EloUpdate}( W, \{v^a_i\}_{i \in I})$ \Comment*[r]{Update agent Elo ratings}
    $\mathcal{A} \gets \mathcal{A} \cup \{(\hat{a}_i, \hat{l}^a_i, \hat{v}^a_i= v^a_i) \mid i \in I, \hat{a}_i \ne a_i\}$\Comment*[r]{Expand agent ensemble}
}
\Return{$\mathcal{S},\mathcal{A}$}
\end{algorithm}

\section{Task}
\label{sec:task}

\subsection{In-Context Operator Networks}

Operator learning is the problem of approximating the map between functions from data. A broad range of the scientific machine learning problems can be formulated as operator learning. For example, problems in weather prediction, fluid dynamics, and molecular simulation can be abstracted as approximating the solution operator of a differential equation that maps an initial condition to the future state.

Standard neural operator methods
learn these maps from input-output function pairs, training one model per operator. In-Context Operator Networks
(ICON)~\citep{yang2023icon} makes a leap forward: a single model receives $k$ input-output example pairs at inference time and infers the
hidden operator on the fly without fine-tuning. Several extensions have advanced the core framework:
ICON-LM~\citep{yang2025iconlm} incorporates an autoregressive
architecture with a next-function prediction training paradigm;
\citet{yang2024icon_pde} demonstrated generalization to new PDE forms
unseen during training; GenICON~\citep{zhang2025genicon} introduced
probabilistic modeling and uncertainty quantification by interpreting ICON
as implicit Bayesian inference. The paradigm has also been applied to
multi-physics fluid dynamics~\citep{cao2024vicon}, graph-structured
spatiotemporal prediction~\citep{wu2026gicon}, and optimal execution in
finance~\citep{meng2025iconfinance,cole2026iconmeasure}. Theoretically,
it is supported by robustness guarantees under domain
shift~\citep{liu2023iconrobust}, generalization
bounds~\citep{cole2024icontheory}, and connections to gradient descent in
function spaces~\citep{mishra2025continuum}.

As an example, we consider ICON for the prediction of 1D systems governed by conservation law~\citep{yang2024icon_pde} with periodic boundary condition:
\begin{equation}\label{eq:conservation_law}
  \partial_t u(t,x) + \partial_x f(u(t,x)) = 0,
  \quad x \in [0,1],
\end{equation}
where $f$ is the flux function. For simplicity, we use $u_t$ to denote the function $u(t,\cdot)$. We can define the forward operator $\mathcal{F}_{f,\tau}\colon u_0 \mapsto u_\tau$, and input and output of ICON model $T_\theta$ (realized as a transformer) write as
\begin{equation}\label{eq:icon}
  \widehat{u}^{(q)}_\tau =
  T_\theta\bigl(\{(u_0^{(i)},u_\tau^{(i)})\}_{i=1}^{k},\,u_0^{(q)}\bigr),
\end{equation}
where each contextual example pair $(u_0^{(i)}, u_\tau^{(i)})$ exemplify the operator $\mathcal{F}_{f,\tau}$ via relationship $u_\tau^{(i)} = \mathcal{F}_{f,\tau}(u_0^{(i)})$. The output $\widehat{u}^{(q)}_\tau$ aims to approximate  
${u}^{(q)}_\tau = \mathcal{F}_{f,\tau} (u_0^{(q)})$ corresponding to the query function $u_0^{(q)}$.

\subsection{Challenge: Example-Count Generalization}
\label{subsec:problem}

We hope to achieve test-time scaling for ICON: more in-context examples should give the model more information about the hidden operator and therefore produce more accurate predictions. In the original ICON architecture~\citep{yang2023icon}, however, this scaling remains an open problem.

The core bottleneck lies in the gap between training constraints and test-time requirements. The model is typically trained with a fixed number of examples (e.g., $k=5$), but must generalize to much larger $k$ at inference. As $k$ increases, the sequence length grows, placing tokens from additional examples at positions the model never encountered during training. In the original design, positions are encoded using a learned embedding table indexed by example slots. This table has a fixed capacity: it lacks entries for any examples beyond the fifth, causing prediction performance to degrade sharply as soon as the sequence exceeds its pre-defined limit.

Redesigning the positional encoding (PE) and model architecture to support variable example counts presents a concrete research problem with a vast design space. Solving it is not a simple parameter tweak; it requires navigating a complex research codebase, combining domain expertise with coordinated engineering changes across multiple files, and validating candidates through a complicated training and evaluation pipeline. We applied EvE to this problem, with the agent ensemble operating directly on the full ICON source repository, a complete research codebase with model implementation, training pipeline, and evaluation infrastructure.

\subsection{Experiment Setup}
\label{subsec:evaluation}

The experiments in this work use the benchmark of 1D conservation law with random cubic flux from~\citet{yang2024icon_pde}:
\begin{equation}\label{eq:weno}
  \partial_t u + \partial_x(au^3 + bu^2 + cu) = 0,
  \qquad x \in [0,1],
\end{equation}
where $a, b, c \sim \mathrm{Uniform}[-1,1]$ are sampled independently for
each operator instance. Each instance maps an initial condition drawn from a
periodic Gaussian random field to the solution at $\tau = 0.1$ (details in
Appendix~\ref{sec:appendix_setup}). The training set contains 1{,}000
operator instances with 100 initial conditions each.

At training time the model sees $k = 5$ in-context examples. At evaluation
it is tested with $k = 1, 2, \ldots, 10$: the first five fall within the
training distribution, while $k = 6$ through $k = 10$ constitute a strict
out-of-distribution regime in which the model must handle longer sequences
than it ever encountered during training.

For each example count $k$, the error is the mean absolute error on a
held-out validation set:
\begin{equation}\label{eq:per_example_error}
  e_k = \frac{1}{|\mathcal{V}|}\sum_{v \in \mathcal{V}}
  \frac{1}{N_x}\sum_{j=1}^{N_x}
  \bigl|\widehat{u}^{(v)}_\tau(x_j) - u^{(v)}_\tau(x_j)\bigr|,
\end{equation}
where $\mathcal{V}$ is the validation set and $N_x = 100$ is the number of
spatial grid points. The headline metric averages over all example counts:
\begin{equation}\label{eq:mean_error}
  \overline{e} = \frac{1}{10}\sum_{k=1}^{10} e_k.
\end{equation}
The solver score used by EvE (corresponding to $v^s_j$ in
Algorithm~\ref{alg:eve-search}) is $s = -\overline{e}$ (higher is better). During the evolutionary search, each candidate is trained for 2{,}000
steps, sufficient to rank PE designs and enable rapid iteration.


\section{Results}
\label{sec:results}

\subsection{EvE Variants}
\label{subsec:experiments}

We design three experimental conditions to isolate the contribution of the
live agent ensemble. Each condition is run twice independently under
identical compute and training budgets:

\begin{itemize}
  \item \textbf{EvE}: the full ensemble condition. The agent population is
    sampled, scored, and updated continuously alongside solver production.
  \item \textbf{Static-Initial}: the initial agent is used throughout the
    entire search, with no agent evolution.
  \item \textbf{Static-Final}: the single best-rated agent from the corresponding completed
    EvE run is extracted and frozen. The same solver search is then run with
    this agent held static.
\end{itemize}

The three conditions form a gradient: no evolution (Static-Initial),
full evolution followed by freezing the best agent (Static-Final), and continuous
evolution (EvE). This isolates whether evolving agents helps, and if so,
whether a single evolved agent suffices or the adaptation must continue
alongside the search. In all runs we set $T=15$ iterations with
$|I|=2$ working agents running in parallel, $|J|=8$ reference solvers,
and $|K|=4$ reference agents per iteration
(Algorithm~\ref{alg:eve-search}). Further configuration details are
provided in Appendix~\ref{app:run_configuration}.
\begin{figure}[!ht]
  \centering
  \includegraphics[width=0.8\linewidth]{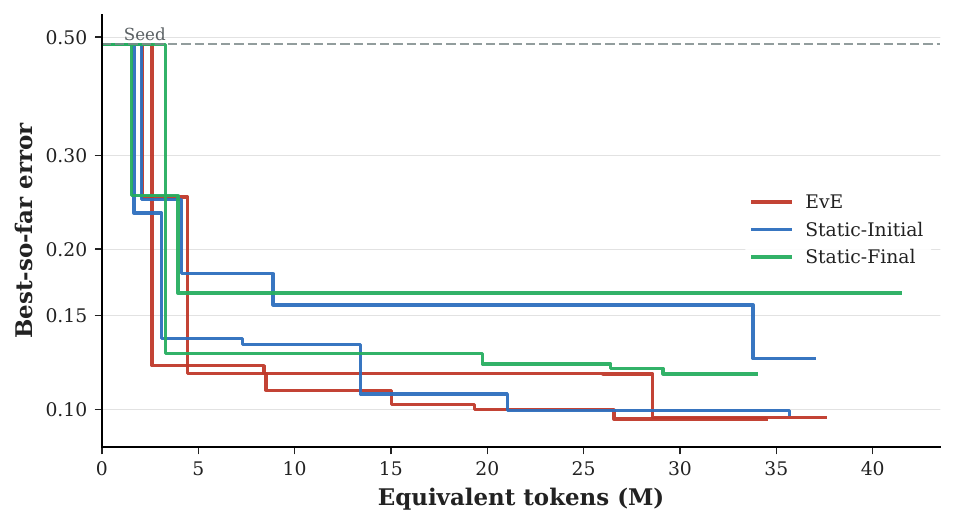}
  \caption{Search trajectories for all three variants (two independent runs
  each). The y-axis is the running minimum of mean error $\overline{e}$
  (lower is better); the x-axis is cumulative equivalent tokens
  $T_{\mathrm{eq}}$ in millions
  (Appendix~\ref{app:compute_normalization}). The gray dashed line marks
  the Seed baseline.}
  \label{fig:live_vs_static}
\end{figure}


Figure~\ref{fig:live_vs_static} compares all three variants. The
two EvE runs (red) descend in near-lockstep, converging to almost identical
final errors. The two Static-Initial runs (blue) diverge: one eventually
approaches EvE, while the other plateaus at a visibly higher level.
One might expect Static-Final to consistently outperform
Static-Initial, since it starts from a higher-rated agent. Instead, one
Static-Final run (green) plateaus early above even the worse Static-Initial
run, while the other lands between the two Static-Initial curves.
The likely explanation is phase mismatch: the frozen agent was optimized for the late
stage of the original EvE run, where the solver population was already
strong, but a fresh search starts from the Seed and requires early-stage
exploration strategies that this agent no longer carries
(Section~\ref{subsec:agent_traces} provides direct evidence for this
phase dependence).

Collectively, these comparisons demonstrate that continuous evolution is indispensable. Eliminating this process, either by omitting it entirely (Static-Initial) or by fixing a specific snapshot (Static-Final), compromises performance and robustness. In Section~\ref{subsec:agent_traces}, we examine the specific agent behaviors that underpin these findings.

\subsection{Example-Count Generalization}
\label{subsec:ood}

\begin{figure}[!ht]
  \centering
  \begin{subfigure}[t]{0.495\linewidth}
    \centering
    \includegraphics[width=\linewidth]{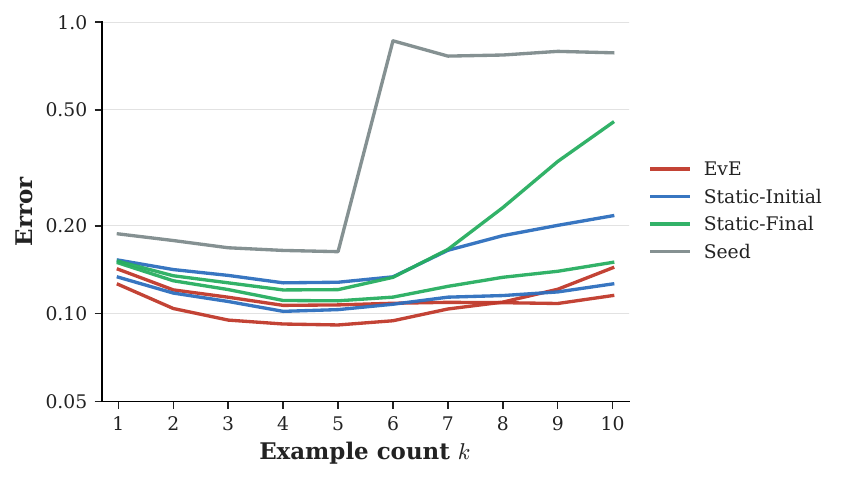}
    \caption{2{,}000 training steps}
    \label{fig:d1_d10_2k}
  \end{subfigure}\hfill
  \begin{subfigure}[t]{0.495\linewidth}
    \centering
    \includegraphics[width=\linewidth]{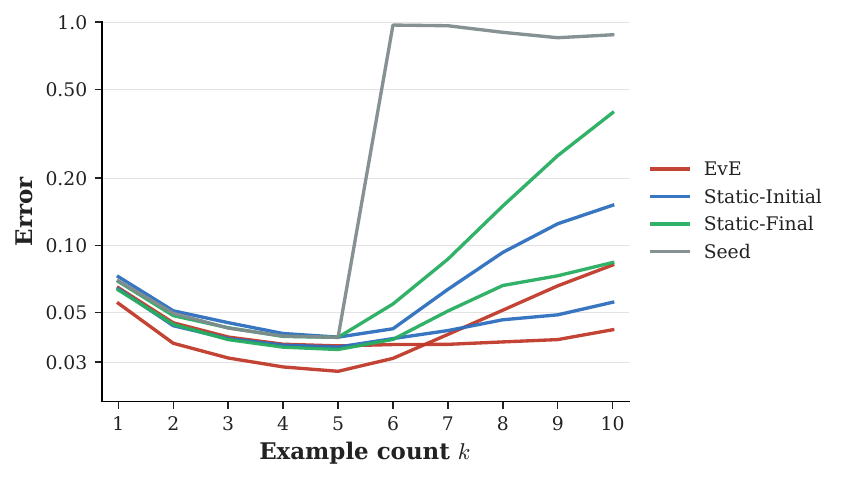}
    \caption{10{,}000 training steps}
    \label{fig:d1_d10_10k}
  \end{subfigure}
  \caption{Per-example-count error curves ($k=1$ through $k=10$) at two
  training budgets. Each variant contributes the best PE method from each of its
  two independent runs; the Seed (gray, ICON vanilla PE) is the reference
  baseline. See Table~\ref{tab:pe_methods} for method details.}
  \label{fig:d1_d10}
\end{figure}

The challenge motivating this work is example-count generalization: whether
the model can handle more in-context examples than it was trained on
(Section~\ref{subsec:problem}). The headline metric $\overline{e}$ averages
over both in-distribution ($k \leq 5$) and out-of-distribution ($k > 5$)
example counts, so it cannot distinguish whether a variant improves within
the trained range or also generalizes beyond it.
Figure~\ref{fig:d1_d10} resolves this by plotting per-example-count error curves for
the best PE method from each of the two independent runs per variant. To investigate the performance of the fully trained model, the top solvers are also evaluated with 10{,}000 training steps for comparison.

The Seed (gray) exposes the baseline failure: error is moderate for
$k \leq 5$ but collapses catastrophically once the example count exceeds
the training boundary. All evolved methods avoid this collapse. Among them,
the EvE (red) performs the best, with error staying below $0.15$ even at
$k=10$ in the 2{,}000-step budget and below $0.08$ at $k=10$ under full training. Both Static-Initial (blue) and Static-Final (green) under-perform the EvE, and demonstrate the lack of robustness.

\subsection{Stage-Dependent Agent Adaptation}
\label{subsec:agent_traces}

Figures~\ref{fig:optimizer_evolution} and~\ref{fig:optimizer_evolution_rep2}
show how agent behavior shifts over the course of each of the two
independent EvE runs.

\begin{figure}[H]
\centering

\definecolor{earlybg}{RGB}{235,245,255}
\definecolor{earlyframe}{RGB}{70,130,180}
\definecolor{midbg}{RGB}{255,248,235}
\definecolor{midframe}{RGB}{180,140,60}
\definecolor{latebg}{RGB}{240,255,240}
\definecolor{lateframe}{RGB}{80,160,80}
\definecolor{codebg}{RGB}{248,248,248}
\definecolor{codeframe}{RGB}{160,160,160}

\begin{minipage}[t]{0.325\textwidth}
\begin{tcolorbox}[
  enhanced, title={\scriptsize (a) Agent guidance update at iteration 1},
  fonttitle=\bfseries\scriptsize,
  colback=earlybg, colframe=earlyframe!80,
  colbacktitle=earlyframe!85, coltitle=white,
  arc=2pt, boxrule=0.4pt,
  top=2pt, bottom=2pt, left=2.5pt, right=2.5pt,
  fontupper=\scriptsize,
  height=7.6cm, valign=top,
]
\raggedright
\textit{``Updated the search directions to make the structured-position family more actionable for ICON specifically. Added a note that the stock \texttt{index\_pos} is block-level rather than token-level, and recorded the slot/role/offset decomposition as a clean diversification probe when examples are dominated by learned-absolute baselines.''}

\vfill
\raggedright\color{earlyframe!80}\scriptsize
$\triangleright$ Identifies flat \texttt{index\_pos} as the bottleneck and steers search toward slot/role/offset decomposition.
\end{tcolorbox}
\end{minipage}%
\hfill
\begin{minipage}[t]{0.325\textwidth}
\begin{tcolorbox}[
  enhanced, title={\scriptsize (b) Agent guidance update at iteration 4},
  fonttitle=\bfseries\scriptsize,
  colback=midbg, colframe=midframe!80,
  colbacktitle=midframe!85, coltitle=white,
  arc=2pt, boxrule=0.4pt,
  top=2pt, bottom=2pt, left=2.5pt, right=2.5pt,
  fontupper=\scriptsize,
  height=7.6cm, valign=top,
]
\raggedright
\textit{``Recorded a workspace-local result: demo-slot position interpolation is currently the strongest visible family in the sampled evidence and should be treated as an active direction rather than a fallback. This is intended to steer later Phase 2 runs away from overcommitting to the copied structured-PE prefill.''}

\vfill
\raggedright\color{midframe!80}\scriptsize
$\triangleright$ Promotes demo-slot interpolation to the active direction and steers working agents to extend it.
\end{tcolorbox}
\end{minipage}%
\hfill
\begin{minipage}[t]{0.325\textwidth}
\begin{tcolorbox}[
  enhanced, title={\scriptsize (c) Agent guidance update at iteration 8},
  fonttitle=\bfseries\scriptsize,
  colback=latebg, colframe=lateframe!80,
  colbacktitle=lateframe!85, coltitle=white,
  arc=2pt, boxrule=0.4pt,
  top=2pt, bottom=2pt, left=2.5pt, right=2.5pt,
  fontupper=\scriptsize,
  height=7.6cm, valign=top,
]
\raggedright
\textit{``Recorded local workspace evidence: interpolation-only variants underperformed the best structured+bias solver, and overflow-triggered hybrids were the unstable family. Added a concrete follow-up suggestion to probe role-conditioned long-context bias compression before revisiting more aggressive overflow hybrids.''}

\vfill
\raggedright\color{lateframe!80}\scriptsize
$\triangleright$ Retracts global rescale and overflow hybrids; redirects search toward conservative long-context compression.
\end{tcolorbox}
\end{minipage}

\medskip

\begin{minipage}[t]{0.325\textwidth}
\begin{tcolorbox}[
  enhanced, title={\scriptsize (d) Solver PE at iter 2 ($\overline{e}=0.117$)},
  fonttitle=\bfseries\scriptsize,
  colback=codebg, colframe=codeframe,
  colbacktitle=codeframe!70, coltitle=white,
  arc=2pt, boxrule=0.4pt,
  top=2pt, bottom=2pt, left=4pt, right=4pt,
  fontupper=\scriptsize,
  height=6.5cm, valign=top,
]
\raggedright
\texttt{InterpolatedDemoPE.forward} (excerpt):
\begin{lstlisting}[basicstyle=\ttfamily\tiny,aboveskip=2pt,belowskip=0pt]
demo_id = index_pos // 3   # slot
role    = index_pos %  3   # role
scaled  = demo_id * (mt / max(1, demo_max))
lo, hi  = scaled.floor(), scaled.ceil()
emb     = lerp(table[lo], table[hi], scaled-lo)
return emb + role_emb(role)
\end{lstlisting}
\vfill
\raggedright\color{codeframe}\scriptsize
Implements the slot/role split from (a), rescaling unseen demo indices into the trained $[0,5]$ range.
\end{tcolorbox}
\end{minipage}%
\hfill
\begin{minipage}[t]{0.325\textwidth}
\begin{tcolorbox}[
  enhanced, title={\scriptsize (e) Solver PE at iter 5 ($\overline{e}=0.186$)},
  fonttitle=\bfseries\scriptsize,
  colback=codebg, colframe=codeframe,
  colbacktitle=codeframe!70, coltitle=white,
  arc=2pt, boxrule=0.4pt,
  top=2pt, bottom=2pt, left=4pt, right=4pt,
  fontupper=\scriptsize,
  height=6.5cm, valign=top,
]
\raggedright
\texttt{OverflowAware\allowbreak{}InterpolatedDemoPE} (excerpt):
\begin{lstlisting}[basicstyle=\ttfamily\tiny,aboveskip=2pt,belowskip=0pt]
emb_base = interp_demo(demo_id)
overflow = relu(demo_id - max_train)
gate     = sigmoid(gate_param)
residual = sin_cos(overflow * freqs)
return emb_base + role_emb(role) +
       gate * overflow_scale * residual
\end{lstlisting}
\vfill
\raggedright\color{codeframe}\scriptsize
Extends (b)'s interpolation backbone with a gated sinusoidal residual beyond the trained demo range.
\end{tcolorbox}
\end{minipage}%
\hfill
\begin{minipage}[t]{0.325\textwidth}
\begin{tcolorbox}[
  enhanced, title={\scriptsize (f) Solver PE at iter 11 ($\overline{e}=0.139$)},
  fonttitle=\bfseries\scriptsize,
  colback=codebg, colframe=codeframe,
  colbacktitle=codeframe!70, coltitle=white,
  arc=2pt, boxrule=0.4pt,
  top=2pt, bottom=2pt, left=4pt, right=4pt,
  fontupper=\scriptsize,
  height=6.5cm, valign=top,
]
\raggedright
\texttt{InterpolatedDemoPE} overflow-only (excerpt):
\begin{lstlisting}[basicstyle=\ttfamily\tiny,aboveskip=2pt,belowskip=0pt]
in_range = clamp(demo_id, max=mt)
overflow = relu(demo_id - mt)
compr    = log1p(overflow / k) * k
coord    = in_range + compr  # no global rescale
emb      = lerp_table(coord)
return emb + role_emb(role)
\end{lstlisting}
\vfill
\raggedright\color{codeframe}\scriptsize
Follows (c): preserves in-range geometry and compresses only the overflow tail with \texttt{log1p}.
\end{tcolorbox}
\end{minipage}

\caption{Run~1: from agent guidance to solver code. \textbf{Top row}:
agent guidance updates appended at iterations 1, 4, and 8.
\textbf{Bottom row}: PE code written by later working agents that read
those updates. Each column traces how a guidance update (top) materializes
as solver code (bottom). Columns (c) and (f) show a late-phase strategy
reversal: the agent retracts an earlier direction, and the matching code
change lands within a few iterations.}
\label{fig:optimizer_evolution}
\end{figure}

\begin{figure}[H]
\centering

\definecolor{earlybg}{RGB}{235,245,255}
\definecolor{earlyframe}{RGB}{70,130,180}
\definecolor{midbg}{RGB}{255,248,235}
\definecolor{midframe}{RGB}{180,140,60}
\definecolor{latebg}{RGB}{240,255,240}
\definecolor{lateframe}{RGB}{80,160,80}
\definecolor{codebg}{RGB}{248,248,248}
\definecolor{codeframe}{RGB}{160,160,160}

\begin{minipage}[t]{0.325\textwidth}
\begin{tcolorbox}[
  enhanced, title={\scriptsize (a) Agent guidance update at iteration 1},
  fonttitle=\bfseries\scriptsize,
  colback=earlybg, colframe=earlyframe!80,
  colbacktitle=earlyframe!85, coltitle=white,
  arc=2pt, boxrule=0.4pt,
  top=2pt, bottom=2pt, left=2.5pt, right=2.5pt,
  fontupper=\scriptsize,
  height=7.6cm, valign=top,
]
\raggedright
\textit{``Added a note under the position-interpolation family: when ICON uses a learned embedding over coarse demo/role IDs, long-context failure may be caused by untrained embedding rows rather than attention behavior. The recommended probe is to decompose the coarse ID into demo-index plus role and interpolate demo coordinates back into the trained range before lookup.''}

\vfill
\raggedright\color{earlyframe!80}\scriptsize
$\triangleright$ Identifies untrained embedding rows and proposes demo/role decomposition with in-range interpolation.
\end{tcolorbox}
\end{minipage}%
\hfill
\begin{minipage}[t]{0.325\textwidth}
\begin{tcolorbox}[
  enhanced, title={\scriptsize (b) Agent guidance update at iteration 5},
  fonttitle=\bfseries\scriptsize,
  colback=midbg, colframe=midframe!80,
  colbacktitle=midframe!85, coltitle=white,
  arc=2pt, boxrule=0.4pt,
  top=2pt, bottom=2pt, left=2.5pt, right=2.5pt,
  fontupper=\scriptsize,
  height=7.6cm, valign=top,
]
\raggedright
\textit{``Recorded a local lesson from this workspace: once a structured/interpolated absolute PE is already in place, a cleaner next probe than adding more absolute gating is to inject demo-distance plus role-pair relative attention bias in the transformer.''}

\vfill
\raggedright\color{midframe!80}\scriptsize
$\triangleright$ Marks gated hybrids as risky and redirects search from absolute-side gating to relative attention bias.
\end{tcolorbox}
\end{minipage}%
\hfill
\begin{minipage}[t]{0.325\textwidth}
\begin{tcolorbox}[
  enhanced, title={\scriptsize (c) Agent guidance update at iteration 14},
  fonttitle=\bfseries\scriptsize,
  colback=latebg, colframe=lateframe!80,
  colbacktitle=lateframe!85, coltitle=white,
  arc=2pt, boxrule=0.4pt,
  top=2pt, bottom=2pt, left=2.5pt, right=2.5pt,
  fontupper=\scriptsize,
  height=7.6cm, valign=top,
]
\raggedright
\textit{``Recorded a concrete follow-up: keep the structured/interpolated PE and structured RoPE path, but change only the relative-bias distance coordinate so the trained demo window stays exact and overflow distance is compressed smoothly instead of globally rescaled.''}

\vfill
\raggedright\color{lateframe!80}\scriptsize
$\triangleright$ Keeps the dominant PE fixed while compressing only overflow distance smoothly.
\end{tcolorbox}
\end{minipage}

\medskip

\begin{minipage}[t]{0.325\textwidth}
\begin{tcolorbox}[
  enhanced, title={\scriptsize (d) Solver PE at iter 1 ($\overline{e}=0.121$)},
  fonttitle=\bfseries\scriptsize,
  colback=codebg, colframe=codeframe,
  colbacktitle=codeframe!70, coltitle=white,
  arc=2pt, boxrule=0.4pt,
  top=2pt, bottom=2pt, left=4pt, right=4pt,
  fontupper=\scriptsize,
  height=6.5cm, valign=top,
]
\raggedright
\texttt{InterpolatedDemo\allowbreak{}RoleEmbedding} (excerpt):
\begin{lstlisting}[basicstyle=\ttfamily\tiny,aboveskip=2pt,belowskip=0pt]
demo_id = index // num_roles
role    = index %  num_roles
denom   = max(demo_id.max(), mt, 1)
scaled  = demo_id * (mt / denom)
lo, hi  = scaled.floor(), clamp(lo+1, mt)
demo_pe = lerp(table[lo], table[hi], scaled-lo)
return demo_pe + role_emb(role)
\end{lstlisting}
\vfill
\raggedright\color{codeframe}\scriptsize
Implements (a)'s in-range interpolation as pure demo/role lookup tables, without local sinusoid.
\end{tcolorbox}
\end{minipage}%
\hfill
\begin{minipage}[t]{0.325\textwidth}
\begin{tcolorbox}[
  enhanced, title={\scriptsize (e) Solver PE at iter 6 ($\overline{e}=0.117$)},
  fonttitle=\bfseries\scriptsize,
  colback=codebg, colframe=codeframe,
  colbacktitle=codeframe!70, coltitle=white,
  arc=2pt, boxrule=0.4pt,
  top=2pt, bottom=2pt, left=4pt, right=4pt,
  fontupper=\scriptsize,
  height=6.5cm, valign=top,
]
\raggedright
\texttt{StructuredFunctionPE} simplified (excerpt):
\begin{lstlisting}[basicstyle=\ttfamily\tiny,aboveskip=2pt,belowskip=0pt]
# absolute side: bare baseline
table = nn.Embedding(mt + 1, dim)
demo_pe = lerp_in_range_table(demo_id)
role_pe = role_emb(role)
local   = scale * sin_cos(offsets)
return demo_pe + role_pe + local
# relative bias in transformer_evolve.py
\end{lstlisting}
\vfill
\raggedright\color{codeframe}\scriptsize
Acts on (b): restores the simplest interpolated absolute PE and moves complexity to relative bias.
\end{tcolorbox}
\end{minipage}%
\hfill
\begin{minipage}[t]{0.325\textwidth}
\begin{tcolorbox}[
  enhanced, title={\scriptsize (f) Solver PE at iter 15 ($\overline{e}=0.108$)},
  fonttitle=\bfseries\scriptsize,
  colback=codebg, colframe=codeframe,
  colbacktitle=codeframe!70, coltitle=white,
  arc=2pt, boxrule=0.4pt,
  top=2pt, bottom=2pt, left=4pt, right=4pt,
  fontupper=\scriptsize,
  height=6.5cm, valign=top,
]
\raggedright
\texttt{StructuredFunctionPE} log1p (excerpt):
\begin{lstlisting}[basicstyle=\ttfamily\tiny,aboveskip=2pt,belowskip=0pt]
in_range = clamp(demo_id, max=mt)
overflow = relu(demo_id - mt)
compr    = k * log1p(overflow / k)
coord    = in_range + compr
scale    = mt / max(coord.max(), mt)
demo_pe  = lerp_table(coord * scale)
return demo_pe + role_emb(role) + local
\end{lstlisting}
\vfill
\raggedright\color{codeframe}\scriptsize
Implements (c): keeps in-range coordinates exact and compresses overflow with \texttt{log1p} before rescaling.
\end{tcolorbox}
\end{minipage}

\caption{Run~2: from agent guidance to solver code, with the same layout as
Figure~\ref{fig:optimizer_evolution}. \textbf{Top row}: agent guidance updates at
iterations 1, 5, and 14. \textbf{Bottom row}: PE code written by working agents.
Column~(a) and~(d) are parallel outputs from the same iteration (the
guidance update and the solver were produced in the same agent session);
columns~(b)$\to$(e) and~(c)$\to$(f) show later working agents reading
accumulated guidance. }
\label{fig:optimizer_evolution_rep2}
\end{figure}

One can clearly see the stage-dependent agent adaptation. In the early iterations (column~a), agents
identify the structural bottleneck and propose broad exploration directions.
In the middle phase (column~b), agents ground their updates in accumulated
solver evidence: they promote the most promising PE family observed so far
and steer subsequent working agents toward extending it. In the late phase
(column~c), agents retract earlier strategies that have stopped producing
gains and redirect search toward finer-grained refinements. The matching
solver-side changes (bottom row) confirm that these agent updates
materialize as concrete code within a few iterations.

The two independent runs exhibit the same progression,
suggesting that this pattern of stage-dependent agent adaptation is a robust property of EvE. Appendix~\ref{app:evolved_pe} describes all six evolved solvers and the
Seed baseline in detail.

\subsection{Computational Cost}
\label{subsec:cost}

All coding-agent sessions were run via Codex using GPT~5.4 at medium
reasoning effort. For agent-side usage, we report the cumulative
equivalent-token metric $T_{\mathrm{eq}}$
(Appendix~\ref{app:compute_normalization}), which aggregates cached input,
fresh input, and output tokens into a single normalized usage scale. This
allows runs with different cache and output profiles to be compared under the
same convention. GPU training and evaluation compute are reported
separately below.

For the GPU usage in model training and evaluation, each iteration evaluates 2 solver candidates in
parallel, each taking no more than 40 minutes on 2 NVIDIA A40 GPUs (typically
around 20 minutes; the variation depends on the computational efficiency
of the candidate's implementation). Each iteration completes in roughly 40 to 60 minutes end-to-end, including the coding-agent session. Over 15
iterations, one complete run takes approximately 10 to 15 hours, consuming 20-30 A40 GPU-hours.

\section{Related Work}
\label{sec:related_work}

\paragraph{LLM-guided code evolution.}
Evolutionary search over executable code has a long history, from genetic
programming \citep{koza1992gp} through AutoML-Zero \citep{real2020automlzero}.
The recent generation of LLM-guided program search
\citep{llm4adsurvey2026,lehman2023elm,liu2024eoh,romeraparedes2024funsearch,novikov2025alphaevolve,algorithmicsuperintelligence2026openevolve,codeevolve,shinkaevolve2025,thetaevolve2025,evox2026,adaevolve2026}
shows that language-model-driven proposals can discover useful scientific and
algorithmic code. Across these systems the evolving object is a single code
artifact and the surrounding evolutionary loop remains a human-designed
scaffold, even when the search strategy itself is
meta-evolved~\citep{evox2026,adaevolve2026}. An orthogonal line replaces the
frozen-LLM evolutionary loop with test-time RL
fine-tuning~\citep{tttdiscover2026}. Classical
coevolution demonstrated that search dynamics can be shaped by interactions
between evolving populations
\citep{hillis1990coevolution,paredis1995coevolutionary,rosin1997competitive}.
Escher-Loop~\citep{escher_loop_2026} introduced this dual-population
structure into LLM-based code search, evaluating guidance states through the
solvers they produce. EvE inherits this credit-assignment mechanism but
departs from the ``LLMs as optimizers'' paradigm: instead of evolving prompt
builders around weaker solvers, EvE evolves agent guidance and skills that steer
already-capable coding agents.

\paragraph{Self-modifying agents.}
A separate line of work evolves the agent itself.
STOP~\citep{stop2024} demonstrated recursive self-improvement of
code-generation scaffolds; G\"{o}del Agent~\citep{godelagent2025} extended
this to full self-referential agent modification. Darwin G\"{o}del
Machine~\citep{dgm2025} scales this idea with an expanding archive of agent variants
selected by empirical fitness. Huxley-G\"{o}del Machine~\citep{hgm2025} refines
this with clade-level credit assignment, evaluating an agent variant by the
aggregate performance of its descendants. HyperAgents~\citep{hyperagents2026}
extends the idea to a fully self-referential program in which the meta-level
modification procedure is itself editable. SICA~\citep{sica2025} removes the
separation between task agent and meta agent entirely, letting a single agent
edit its own code. These systems share a common design choice: the base agent
substrate is part of what evolves. EvE takes the opposite position, fixing the
base agent substrate and restricting evolution to the agent guidance and skills that
organize each agent's behavior.

\paragraph{Skill and guidance evolution.}
Rather than modifying the agent, a growing body of work evolves the skills,
prompts, or guidance that agents receive. OPRO~\citep{opro2024} showed that
language models can optimize prompts from scored histories, and
PromptBreeder~\citep{promptbreeder2023} extended this by evolving the mutation
prompts themselves. More recently, CoEvoSkills~\citep{coevoskills2026}
co-evolves structured skill packages alongside surrogate verifiers as two
coupled populations. Group-Evolving Agents~\citep{gea2026} and
TerraLingua~\citep{terralingua2026} treat the group or ecology as the
evolutionary unit, propagating knowledge through shared experience or
persistent artifacts.
CORAL~\citep{coral2026} combines persistent shared memory with asynchronous
multi-agent exploration for open-ended discovery. EvE belongs to this family:
it fixes the coding agent and evolves agent guidance and skills as a scored
population. The distinguishing mechanism is dual-population credit assignment,
where agent fitness is determined solely by the downstream improvement of
the solvers they help produce, rather than by intrinsic skill quality or
surrogate verification.

\section{Conclusion}

In this work, we introduced EvE, a decentralized framework for algorithmic discovery that organizes highly capable coding agents into a co-evolving system. Rather than reinventing the wheel within the ``LLMs as optimizers'' paradigm, EvE fixes the base agent substrate and focuses entirely on evolving the cumulative guidance and skills that dictate agent behaviors. By maintaining two co-evolving populations, namely functional code solvers and agent guidance states, EvE elegantly integrates solver refinement and self-referential agent optimization into a single, highly parallelized stage.

We demonstrated the efficacy of EvE on a challenging research problem: the positional-encoding bottleneck for example-count generalization in In-Context Operator Networks (ICON). EvE autonomously discovered a robust rescale-then-interpolate mechanism that significantly surpassed the original baseline design. Crucially, our controlled ablations revealed the absolute necessity of stage-dependent agent adaptation. Compared to variants constrained by a fixed initial agent or even a frozen ``best-evolved'' agent, the full EvE framework uniquely avoided early stagnation and phase mismatch. These findings confirm that organizing agents into a live, self-revising ensemble is the fundamental driver for breaking through static performance ceilings. 

Looking forward, EvE's role-free design achieves universal compatibility, establishing a highly flexible foundation for agentic discovery. This architecture naturally supports recursive nesting, allowing any existing multi-agent system, or even an entire ensemble itself, to be seamlessly encapsulated as a single individual within the evolutionary loop. We believe that such self-revising, decentralized ensembles will play a pivotal role in scaling robust automated scientific research.

The evolutionary ensemble perspective also introduces a significant frontier for future inquiry: the optimization of inter-agent connection topology. Much like the interactions in an Ising model, where global order emerges from local spin alignments, the efficacy of an ensemble depends on the precise ``coupling'' between its constituent agents. Future research should address how these connections, which govern information flow and cognitive alignment, can be properly configured. Optimizing these topologies will be essential to ensure that decentralized ensembles do not descend into stochastic noise or, conversely, a collapsed state of perfect alignment where the loss of diversity stifles true discovery. The goal is instead to achieve a phase transition into coherent large-scale scientific reasoning.


\section*{Acknowledgements}
Liu Yang acknowledges support from the National Research Foundation, Singapore, under the NRF fellowship (Project No. NRF-NRFF17-2025-0006). We acknowledge NUS IT's Research Computing group for providing computational support.

\bibliographystyle{plainnat}
\bibliography{references}

\clearpage
\appendix
\section{Experimental Details}
\label{sec:appendix_setup}

This appendix provides the implementation details needed to reproduce the
ablation and interpret the figures. 


\subsection{Compute Normalization}
\label{app:compute_normalization}

Figure~\ref{fig:live_vs_static} uses cumulative equivalent
tokens $T_{\mathrm{eq}}$ as the x-axis. Each working-agent session
records three token categories: cached input tokens $T_{\mathrm{cache}}$
(context reused across turns via prompt caching), fresh input tokens
$T_{\mathrm{fresh}}$ (new content per turn), and output tokens
$T_{\mathrm{out}}$. We normalize these into a single equivalent-token
measure:
\begin{equation}\label{eq:equiv_tokens}
  T_{\mathrm{eq}} = T_{\mathrm{cache}}
    + 2\,T_{\mathrm{fresh}}
    + 12\,T_{\mathrm{out}},
\end{equation}
where the weights 1\,:\,2\,:\,12 follow the relative list-price ratio of
cached input, fresh input, and output tokens for GPT~5.4 at medium reasoning
effort, the fixed model setting used in all coding-agent sessions via Codex.
This normalization maps heterogeneous token types to a single usage scale for
cross-run comparison.

For each iteration $t$ with working-agent set $W_t$,
\[
  \mathrm{step}_{T_{\mathrm{eq}},t} = \sum_{w \in W_t} T_{\mathrm{eq}}(w),
  \qquad
  \mathrm{cumulative}_{T_{\mathrm{eq}},t}
    = \sum_{\tau \le t} \mathrm{step}_{T_{\mathrm{eq}},\tau}.
\]

\paragraph{Cache efficiency.}
Across all six ablation runs, cached tokens account for
94.1\% of total input tokens. This high cache-hit rate arises because
coding-agent sessions maintain a persistent conversation context: repository
files, task instructions, and prior tool-call outputs are re-sent on every
turn, hitting the prompt cache. Each working agent averages 13 turns per
session, so the effective fresh-input fraction is only 5.9\%. The total
equivalent-token budget per run is 34--42M $T_{\mathrm{eq}}$, corresponding
to approximately 186M raw tokens but only 13M tokens of novel content.

This makes the equivalent-token cost of the entire ablation comparable to a
single large-scale training job. GPU training and remote evaluation costs are
not included in $T_{\mathrm{eq}}$.

\subsection{Run Configuration}
\label{app:run_configuration}

\begin{center}
\small
\begin{tabular}{@{}p{0.36\linewidth}p{0.56\linewidth}@{}}
\toprule
Item & Setting \\
\midrule
Task & ICON PE design on the conservation-law benchmark \\
Coding-agent model & GPT~5.4 (medium reasoning effort) via Codex \\
Variants & EvE, Static-Initial, Static-Final \\
Independent runs & 2 per variant \\
Total iterations $T$ & 15 \\
Working agents $|I|$ & 2 per iteration (parallel) \\
Reference solvers $|J|$ & 8 per iteration (rank-biased sampling) \\
Reference agents $|K|$ & 4 per iteration (rank-biased sampling) \\
Solver training & 2{,}000 steps, 2-GPU DDP, bf16 \\
Agent ratings & Elo-style relative ratings (see below) \\
\bottomrule
\end{tabular}
\end{center}

\paragraph{Elo rating system.}
Agents are rated through pairwise comparison of the solvers they produce
within each iteration. Each agent starts with a rating of 1500. After an
iteration, the agent whose solver achieved lower error wins the comparison.
Ratings are updated using the standard Elo formula with update factor $K=32$:
if the expected win probability for agent $i$ is
$E_i = 1/(1+10^{(R_j - R_i)/400})$, then the rating update is
$R_i \gets R_i + K \cdot (S_i - E_i)$, where $S_i \in \{0, 0.5, 1\}$ is the
match outcome. Higher-rated agents are sampled more frequently via
rank-biased Softmax selection.

Each run consists of a seed evaluation (iteration~0) plus 15 evolutionary
iterations with 2 working agents each, yielding a maximum of 31 solvers per
run. EvE produced 31 and 30 valid solvers across its two runs (the single
missing solver is due to a boundary-check violation at iteration~7 of
run~2). Static-Initial and Static-Final each produced 31 valid solvers per
run.

\subsection{Working-Agent Interface}
\label{app:worker_interface}

Each working agent $a_i$ from Algorithm~\ref{alg:eve-search} is
instantiated as a single working-agent session.
Section~\ref{subsec:closed_loop_search} describes the workspace at the
algorithmic level; for reproduction, the important interface facts are the
editable surface, the validation contract, and the score semantics.

Each working agent starts from a repository snapshot plus a prefilled
positional-encoding candidate. It may edit exactly five Python/YAML files:
\begin{itemize}\setlength{\itemsep}{1pt}
  \item \path{configs/experiment/evolve_base.yaml}
  \item \path{configs/model/icon_evolve.yaml}
  \item \path{src/models/icon/icon_evolve.py}
  \item \path{src/models/base/transformer_evolve.py}
  \item \path{src/models/icon/pe_evolve.py}
\end{itemize}
These files shadow the vanilla ICON model and define the complete solver
submission surface for the ablation.

Before stopping, the working agent must invoke the configured \texttt{check-runner}
validation contract. The check verifies that edits stay inside the permitted
surface, that the modified code imports, and that the candidate can run the
target smoke evaluation. Candidates that violate the boundary are not counted
as valid solvers. Valid candidates are then scored by the ICON evaluator using
Equation~\ref{eq:mean_error}; the solver score stored by EvE is
$s=-\overline{e}$, so higher is better.

The working agent also leaves process evidence. Session notes and evaluation
artifacts are preserved with the produced solver and, when the guidance tree
is revised, with the resulting agent entry. These logs are the concrete
source of the cumulative working logs $l^a_i$ in
Equation~\ref{eq:eve_worker_call}.

\subsection{Seed Guidance}
\label{app:seed_guidance}

All runs start from the same seed guidance. The agent's guidance is stored as
a file tree (the \emph{guidance tree}) containing four documents and one
skill:
\begin{itemize}\setlength{\itemsep}{1pt}
  \item \texttt{problem.md}
  \item \texttt{directions.md}
  \item \texttt{mutation\_surface.md}
  \item \texttt{literature\_notes.md}
  \item \texttt{skills/read-eval/SKILL.md}
\end{itemize}
In the EvE variant, this tree is updated by the agent population as the
search progresses; in the Static-Initial variant, it remains unchanged
throughout.

\subsubsection*{Task Context}

\begin{promptbox}[\texttt{problem.md}: Task context provided to each working agent]
ICON is a transformer-based in-context operator network trained on the
conservation-law benchmark. In this example, training uses \texttt{num\_examples=5}, then
evaluation measures error across \texttt{d1..d10} example counts. The practical
goal is simple: find a positional encoding change that helps the model
generalize across example count without damaging the short-context regime.

\medskip
The solver score is the negative of \texttt{mean\_d1\_d10}, so lower average
error on \texttt{d1..d10} means a better score. Keep an eye on
\texttt{mean\_d1\_d4} and the full \texttt{d1..d10} curve as diagnostics: a
candidate that improves the average by collapsing the short-context regime is
not a satisfying result.

\medskip
The training budget is intentionally modest: 2k steps, 2-GPU DDP, bf16 mixed
precision, roughly one iteration-scale experiment.
\end{promptbox}

\subsubsection*{Positional-Encoding Search Menu}

\begin{promptbox}[\texttt{directions.md}: Seed search menu (abbreviated)]
Eight positional-encoding families are presented as a structured search menu.
Each entry gives a short description, and several entries point to
representative references or ICON-specific adaptations.

\medskip
\begin{enumerate}\setlength{\itemsep}{2pt}
  \item \textbf{Sinusoidal absolute}: fixed sinusoidal bases,
    extrapolation-friendly.
  \item \textbf{Learned absolute}: per-position learned embeddings, vanilla
    baseline.
  \item \textbf{RoPE family}: relative phase via Q/K rotations;
    NTK-aware and YaRN variants for interpolation.
  \item \textbf{ALiBi family}: distance-dependent attention bias, no
    explicit position vectors.
  \item \textbf{Relative position bias}: pairwise distance bias in the
    attention matrix.
  \item \textbf{Position interpolation / context-length scaling}: map
    long-context positions back into the trained coordinate range.
  \item \textbf{Structured / hierarchical decomposition}: multi-axis
    encoding separating example identity from within-example token role.
  \item \textbf{Hybrid / gated mechanisms}: combine two or more PE signals
    with learned gating.
\end{enumerate}

\medskip
\emph{Meta-rule:} The family list is a starting map, not a closed set. If the
running best has stalled, the search has narrowed, not exhausted the surface.
Workers are encouraged to explore non-dominant families and cite literature
when introducing new mechanisms.
\end{promptbox}

\subsubsection*{Editable Files}

\begin{promptbox}[\texttt{mutation\_surface.md}: Editable file boundaries]
Five files may be edited, starting as no-op shadows of vanilla ICON:
\begin{itemize}\setlength{\itemsep}{1pt}
  \item \texttt{configs/experiment/evolve\_base.yaml}: dataset and runtime
    defaults.
  \item \texttt{configs/model/icon\_evolve.yaml}: Hydra entrypoint for model
    composition.
  \item \texttt{src/models/icon/icon\_evolve.py}: top-level model hook.
  \item \texttt{src/models/base/transformer\_evolve.py}: attention-side PE
    changes.
  \item \path{src/models/icon/pe_evolve.py}: scratch space for new
    positional modules.
\end{itemize}
\end{promptbox}

\subsubsection*{Persistent Literature Notes}

\begin{promptbox}[\texttt{literature\_notes.md}: Accumulated research notes]
This document starts empty and serves as persistent memory for literature
findings. Workers are instructed to append method names, sources, and short
applicability notes when external positional-encoding references inform a
guidance update.
\end{promptbox}

\subsubsection*{Evaluation-Reading Skill}

\begin{promptbox}[\texttt{skills/read-eval/SKILL.md}: Guidance-local skill]
The seed guidance exposes a small skill for interpreting ICON PE evaluations.
It tells the working agent to read score cards and evaluation summaries, inspect both
\texttt{mean\_d1\_d10} and \texttt{mean\_d1\_d4}, distinguish broad curve
improvements from single-count wins, and separate visible evidence from
speculation when writing future guidance.
\end{promptbox}

\clearpage
\subsection{Evolved Positional-Encoding Methods}
\label{app:evolved_pe}

The evolutionary search produced six evolved solvers across five distinct PE class names from
the three ablation variants, plus the ICON vanilla baseline.  We keep the
logged PE class name as the row label because it is the stable identifier
recorded in the run artifacts, but several winning solvers also revise the
attention-side position mechanism.  Table~\ref{tab:pe_methods} summarizes the
combined position-handling design and compares performance at 2{,}000 and
10{,}000 training steps under identical conditions.

\begin{table}[!htbp]
\centering
\footnotesize
\setlength{\tabcolsep}{3.5pt}
\resizebox{\linewidth}{!}{%
\begin{tabular}{@{}ll l l r r@{}}
\toprule
Variant & Run & PE Class & Key Mechanism
  & $\overline{e}_{2\text{k}}$ & $\overline{e}_{10\text{k}}$ \\
\midrule
EvE & 1 & \texttt{InterpolatedDemoPE}
  & Interpolated demo PE + structured demo/role attention bias
  & 0.114 & \textbf{0.041} \\
EvE & 2 & \texttt{StructuredFunctionPE}
  & Interpolated demo PE + local sinusoid + structured RoPE/bias
  & 0.108 & 0.045 \\
\addlinespace
Static-Initial & 1 & \texttt{StructuredDemoRolePE}
  & Power-law demo compression + sinusoid + demo-aware ALiBi/RoPE
  & 0.115 & 0.045 \\
Static-Initial & 2 & \texttt{RoleOnlyPE}
  & Role-only PE + compressed-demo rotary attention
  & 0.159 & 0.072 \\
\addlinespace
Static-Final & 1 & \texttt{StructuredInterpolationPE}
  & Linear clamp + sinusoidal example signal
  & 0.197 & 0.117 \\
Static-Final & 2 & \texttt{StructuredDemoRolePE}
  & Tanh demo compression + structured RoPE/bias
  & 0.128 & 0.053 \\
\addlinespace
Seed & n/a & \texttt{VanillaICON} (\texttt{nn.Embedding})
  & Flat learned lookup (no structure)
  & 0.485 & 0.480 \\
\bottomrule
\end{tabular}}
\caption{Evolved PE methods and their performance under standardized
retraining at 2{,}000 and 10{,}000 steps.  These scores differ from the
search-time scores in the per-iteration tables
(Appendix~\ref{app:per_iteration}) because standardized retraining uses a
fresh random seed and fixed hyperparameters, whereas search-time evaluation
is conducted within the iteratively constructed solver snapshot.  Rows are
indexed by the logged PE class name, while the Key Mechanism column
summarizes any paired attention-side change.  All evolved methods
improve from 2k to 10k; Static-Final~run~1 remains the weakest transfer because
its high-example-count tail stays large.}
\label{tab:pe_methods}
\end{table}

All six evolved solvers exploit the role/example decomposition of the
flat token-position index $p$ into a \emph{within-example role}
$r = p \bmod 3$ (Key, Value, or Query) and an \emph{example index}
$m = \lfloor p/3 \rfloor$ somewhere in the position-handling stack.
Five encode the example axis directly in the additive PE; \texttt{RoleOnlyPE}
removes it from the additive PE but reintroduces it through demo-aware rotary
attention.  The methods diverge in how they compress overflow demos and in
whether they also add structured attention bias.
We describe each method below, starting from the vanilla baseline.

\subsubsection*{Seed: ICON Vanilla PE (\texttt{nn.Embedding})}

The default ICON positional encoding is a standard learned embedding table:
\[
  \text{PE}(p) = \mathbf{E}[p], \qquad \mathbf{E} \in \mathbb{R}^{N \times d},
\]
where $N = 100$ is a fixed table size and $p$ is the flat integer position
index.  During training with $k=5$ examples, the largest example index
$m_{\text{train}}$ takes the value 5 in the rescaling formulas below, and
the flat position indices range from 0 to 17 (each of the six slots,
five demos plus the query, is allocated three indices to maintain a
consistent $p \bmod 3$ role decomposition, though two of the 18 indices
are structurally unused at runtime due to the training configuration,
and the query output block is the prediction target rather than an input,
so 16 positions carry input tokens despite the $3k+2=17$ semantic block
count in Section~\ref{subsec:problem}).  At test time with $k > 5$, each additional example introduces new
position indices beyond this range, corresponding to untrained embedding
rows and producing essentially random vectors.  This explains the dramatic OOD collapse visible
in Figure~\ref{fig:d1_d10}: error jumps from
$\sim$0.05 (in-distribution) to $\sim$0.9 (out-of-distribution), an
18$\times$ degradation.  The Seed demonstrates that positional-encoding
design, not model capacity, is the bottleneck for example-count generalization.

\subsubsection*{EvE Run 1: InterpolatedDemoPE}

\textbf{Agent:} evolved. The same PE family first appeared at iteration~2,
but the best solver using this PE design was produced at iteration~15.
\textbf{Best at:} iteration~15 ($\overline{e} = 0.096$).

This method decomposes the flat index into role and example axes, then
\emph{globally rescales} all example indices into the training range:
\[
  r = p \bmod 3, \qquad
  m = \lfloor p / 3 \rfloor, \qquad
  \tilde{m} = m \cdot \frac{m_{\text{train}}}
    {\max(m_{\max}, m_{\text{train}})},
\]
where $m_{\text{train}} = 5$ is the maximum example index seen during training
and $m_{\max}$ is the largest example index in the current batch.  The rescaled
coordinate $\tilde{m}$ is guaranteed to lie in $[0, m_{\text{train}}]$ for
any test-time example count.

The positional signal is computed by interpolating between adjacent rows of
a small learned embedding table $\mathbf{D} \in \mathbb{R}^{(m_{\text{train}}+1) \times d}$:
\[
  \text{PE}(p) = \mathbf{R}[r]
    + (1 - f) \cdot \mathbf{D}[\lfloor \tilde{m} \rfloor]
    + f \cdot \mathbf{D}[\lceil \tilde{m} \rceil],
\]
where $f = \tilde{m} - \lfloor \tilde{m} \rfloor$ is the fractional part and
$\mathbf{R} \in \mathbb{R}^{3 \times d}$ is a learned role embedding.  The
key insight is that role identity is preserved exactly (discrete lookup) while
the example axis is continuously compressed, allowing smooth generalization to
unseen example counts.

The published solver is not PE-only.  Its transformer also replaces vanilla
self-attention with a structured additive bias over rescaled demo distance,
same-demo matches, and role pairs.  In other words, both the additive PE and
the attention logits use the same demo/role decomposition.  On the additive-PE
side, the new parameters are still small: two embedding tables
($3d + 6d = 9d$), plus a lightweight set of attention-bias scalars and
role-pair residuals in the evolved encoder layer.

\subsubsection*{EvE Run 2: StructuredFunctionPE}

\textbf{Agent:} evolved (post-seed, suggested ``adding a relative
attention bias over example distance''). \textbf{Best at:} iteration~11
($\overline{e} = 0.097$).

This solver keeps the same demo-axis interpolation as EvE~run~1 but changes
both the additive PE and the attention path.  The additive PE is
\[
  \text{PE}(p) = \mathbf{D}[\tilde{m}]_{\text{interp}}
    + \mathbf{T}[r]
    + \lambda \cdot \text{sinusoid}(o(p)),
\]
where $\mathbf{T} \in \mathbb{R}^{3 \times d}$ is a token-type embedding,
$\lambda$ is a learned scalar, and $o(p)$ is the local offset assigned to
repeated occurrences of the same flat index.  The example index is rescaled
identically to EvE~run~1:
$\tilde{m} = m \cdot m_{\text{train}} / \max(m_{\max}, m_{\text{train}})$.

The transformer then threads the same structured coordinates through
self-attention: it applies RoPE separately on the demo-id and local-offset
axes, and adds a demo-distance bias over role pairs.  Despite being produced
by a different agent than EvE~run~1, this solver independently converged to
the same rescale-then-interpolate mechanism for the demo axis, providing
evidence that the evolutionary search robustly discovers this solution.

\subsubsection*{Static-Initial Run 1: StructuredDemoRolePE}

\textbf{Agent:} seed (fixed, no evolution). \textbf{Best at:}
iteration~15 ($\overline{e} = 0.098$).

Without evolved guidance, this method uses a \emph{sinusoidal} encoding for
the example axis (rather than learned interpolation) combined with a nonlinear
overflow compression:
\[
  \tilde{m} =
  \begin{cases}
    m & \text{if } m \leq m_{\text{train}} \\
    m_{\text{train}} + \alpha \cdot (m - m_{\text{train}})^{0.5}
      & \text{otherwise}
  \end{cases}
\]
where $\alpha$ is the \texttt{extrapolation\_scale} parameter.  The
square-root compression ensures that overflow positions grow sublinearly,
staying close to the training boundary.  The positional signal is:
\[
  \text{PE}(p) = g_d \cdot \text{sinusoid}(\tilde{m}) + g_r \cdot \mathbf{R}[r],
\]
where $g_d, g_r$ are learned scalar gates that balance the example and role
components.  This solver also revises the attention-side position mechanism,
pairing the same demo/role coordinates with ALiBi-style distance bias and
partial structured RoPE.  It achieves strong 10k performance (0.045),
competitive with EvE, showing that a sinusoidal demo axis can remain
effective when both the additive PE and the attention path share the same
compressed structure.

\subsubsection*{Static-Initial Run 2: RoleOnlyPE}

\textbf{Agent:} seed (fixed, no evolution). \textbf{Best at:}
iteration~13 ($\overline{e} = 0.125$).

This is the most aggressive simplification of the additive PE:
\[
  \text{PE}(p) = \lambda \cdot \mathbf{R}[p \bmod 3],
\]
where $\lambda$ is a single learned scalar.  With only 3 embedding rows and
one scalar, this PE has the fewest learnable parameters of any method ($3d + 1$).

The example axis is not removed from the solver entirely, however.  The custom
attention layer still uses compressed demo coordinates
\[
  \tilde{m} =
  \begin{cases}
    m & \text{if } m \leq m_{\text{train}} \\
    m_{\text{train}} + \alpha \log\!\Bigl(1 + \tfrac{m - m_{\text{train}}}{\alpha}\Bigr)
      & \text{otherwise}
  \end{cases}
\]
inside full rotary attention, with the scalar position formed from both
$\tilde{m}$ and the role index $r$.  This design therefore removes example
structure from the additive PE while keeping it in the attention coordinates.
It works surprisingly well in-distribution but shows the widest gap between 2k
(0.159) and 10k (0.072) performance among the successful methods, suggesting
that longer training partially compensates for the weaker explicit PE signal
but cannot fully replace a richer demo-aware position map.

\subsubsection*{Static-Final Run 1: StructuredInterpolationPE}

\textbf{Agent:} best-evolved agent from EvE~run~1, then frozen.
\textbf{Best at:}
iteration~2 ($\overline{e} = 0.165$ during search).

This method uses fixed-frequency sinusoidal encoding with linear overflow
clamping:
\[
  \tilde{m} =
  \begin{cases}
    m & \text{if } m \leq m_{\text{train}} \\
    m_{\text{train}} + 0.5 \cdot (m - m_{\text{train}})
      & \text{otherwise}
  \end{cases}
\]
\[
  \text{PE}(p) = \text{sinusoid}_{\text{fixed}}(\tilde{m}) + \sigma \cdot \mathbf{R}[r],
\]
where the sinusoidal frequencies are precomputed buffers (not learned) and
$\sigma$ is the \texttt{role\_scale} parameter.  Unlike EvE's learned
interpolation, the fixed frequencies cannot adapt their spectral content to
the test-time position range.  In contrast to the other successful methods,
the transformer stays on the vanilla attention path here: the mutation lives
entirely in the additive PE.

Under standardized retraining, this method is trainable and improves with
longer training, but it remains the weakest evolved transfer because the
linear overflow rule leaves a large high-$k$ tail (see case study below).

\subsubsection*{Static-Final Run 2: StructuredDemoRolePE}

\textbf{Agent:} best-evolved agent from EvE~run~2, then frozen.
\textbf{Best at:}
iteration~12 ($\overline{e} = 0.117$).

The agent guidance for this run explicitly noted: ``decompose the ID into
example-index and role, then interpolate example-index coordinates back into the
trained range.''  The resulting PE uses bounded tanh compression:
\[
  \tilde{m}_{\mathrm{PE}} =
  \begin{cases}
    m & \text{if } m \leq m_{\text{train}} - 1 \\
    m_{\text{train}} - 1 + \Delta \cdot \tanh\!\bigl(\tfrac{m - (m_{\text{train}} - 1)}{\tau}\bigr)
      & \text{otherwise}
  \end{cases}
\]
where $\Delta$ is the overflow span and $\tau$ is an overflow temperature.
The tanh mapping guarantees that overflow positions are bounded within
$[m_{\text{train}} - 1,\; m_{\text{train}} - 1 + \Delta)$.  In the actual
solver, the furthest demo in the current sequence is then explicitly anchored
back to $m_{\text{train}}$, preventing the last OOD slot from drifting past
the trained boundary.  The positional
signal is:
\[
  \text{PE}(p) = g_d \cdot \text{sinusoid}(\tilde{m}_{\mathrm{PE}})
    + g_r \cdot \mathbf{R}[r],
\]
with learned scalar gates $g_d, g_r$.  The transformer uses the same
tanh-compressed demo/role coordinates inside both structured RoPE and a
demo-distance role-pair bias.  This method achieves solid 10k performance
(0.053), demonstrating that frozen guidance, when it carries the right
structural insight, can produce competitive designs even without live
adaptation.

\paragraph{Case study: Static-Final run 1 transfer degradation.}
Static-Final~run~1 is the weakest evolved method when transferred to the standardized
training pipeline, but it shows degradation rather than total collapse.
Its error is 0.197 at step~2{,}000 and 0.117 at step~10{,}000: longer
training repairs the in-distribution regime, but the tail remains large,
reaching 0.393 at $k=10$.  In contrast, the same PE scored 0.165 during the
original evolutionary search (the worst evolved score, but still far better
than the baseline's 0.485).

Three factors converge to explain this degradation:

\begin{enumerate}
\item \textbf{Weak overflow compression.}  The PE clips overflow example
  indices with a fixed linear scale factor (\texttt{extrapolation\_scale=0.5}).
  This keeps positions closer to the training boundary than the vanilla
  embedding table, but it still lets large example counts drift farther than
  the learned-interpolation and tanh-compression designs.

\item \textbf{Fixed spectral signal.}  The example axis is encoded by
  sinusoidal frequencies stored as buffers rather than learned interpolation
  rows.  The signal is useful enough to train down the short-context region,
  but it cannot reshape itself around the OOD portion of the curve.

\item \textbf{Context mismatch.}  The frozen agent was originally the
  best agent from EvE~run~1, where it operated within a diverse,
  evolving population.  When frozen into Static-Final's static environment,
  its guidance could not adapt to the solvers it was now directing. The PE
  therefore preserved a locally useful structured sinusoidal design, but never
  received the later pressure that pushed the live EvE runs toward stronger
  learned interpolation or bounded overflow compression.
\end{enumerate}

This case illustrates a key advantage of the live ensemble: agent
adaptation keeps pressure on the unresolved tail across iterations, whereas
frozen guidance cannot revise itself in response to the curve it is now
producing.

\subsection{Per-Iteration Results}
\label{app:per_iteration}

The tables below show per-iteration results for all six runs, including
cumulative equivalent tokens $T_{\mathrm{eq}}$ (Equation~\ref{eq:equiv_tokens}).
Tables are ordered \textbf{EvE}, \textbf{Static-Initial}, \textbf{Static-Final}, with the two
independent runs of each variant on adjacent pages.
All six tables share the same five-column schema; the \texttt{PE Class}
column tracks architecture choice per iteration (see
Section~\ref{app:evolved_pe} and Table~\ref{tab:pe_methods} for the
definition of each logged PE class name). Some winning solvers also revise
the attention-side position mechanism; Section~\ref{app:evolved_pe}
summarizes that full solver surface (PE classes that appear only in early
iterations and are superseded, such as \texttt{InterpolatedDemoRoleEmbedding}
in EvE run~2 iteration~1, are not described separately). For EvE runs this
column varies as the agent evolves; for Static-Initial and Static-Final runs the fixed agent consistently
produces the same PE class, making the contrast between evolved and frozen
PE labels directly visible.
Each iteration nominally uses two independent working agents; the
$\overline{e}$ column reports the best of the two. Bold entries in the
best-so-far column mark iterations where the running minimum was matched
or improved. Iterations where a working agent failed are footnoted under
the relevant table.

\newcommand{\perIterCaption}[1]{\footnotesize\textbf{Table~\thetable.}\ #1\par\vspace{2pt}}

\refstepcounter{table}\label{tab:eve1_per_iter}%
\begin{center}
\begin{minipage}{\linewidth}\centering
\perIterCaption{EvE run 1 per-iteration results.}
{\scriptsize
\begin{tabular}{@{}rlrrr@{}}
\toprule
Iter & PE Class & $\overline{e}$ & Best-so-far & $T_{\mathrm{eq}}$ (M) \\
\midrule
0 & \texttt{VanillaICON} & 0.4848 & 0.4848 & 0.0 \\
1 & \texttt{StructuredFunctionPE} & 0.2504 & \textbf{0.2504} & 2.1 \\
2 & \texttt{InterpolatedDemoPE} & 0.1169 & \textbf{0.1169} & 4.5 \\
3 & \texttt{StructuredFunctionPE} & 0.2211 & 0.1169 & 6.3 \\
4 & \texttt{InterpolatedDemoPE} & 0.1086 & \textbf{0.1086} & 8.5 \\
5 & \texttt{InterpolatedDemoPE} & 0.1142 & 0.1086 & 11.1 \\
6 & \texttt{InterpolatedDemoPE} & 0.1021 & \textbf{0.1021} & 15.0 \\
7 & \texttt{InterpolatedDemoPE} & 0.1021 & \textbf{0.1021} & 17.0 \\
8 & \texttt{InterpolatedDemoPE} & 0.0999 & \textbf{0.0999} & 19.4 \\
9 & \texttt{InterpolatedDemoPE} & 0.1008 & 0.0999 & 23.2 \\
10 & \texttt{InterpolatedDemoPE} & 0.1009 & 0.0999 & 25.1 \\
11 & \texttt{InterpolatedDemoPE} & 0.0959 & \textbf{0.0959} & 26.6 \\
12 & \texttt{InterpolatedDemoPE} & 0.0959 & \textbf{0.0959} & 28.3 \\
13 & \texttt{InterpolatedDemoPE} & 0.0959 & \textbf{0.0959} & 30.0 \\
14 & \texttt{InterpolatedDemoPE} & 0.0959 & \textbf{0.0959} & 32.1 \\
15 & \texttt{InterpolatedDemoPE} & 0.0957 & \textbf{0.0957} & 34.5 \\
\bottomrule
\end{tabular}\par
}
\end{minipage}
\end{center}

\medskip

\refstepcounter{table}\label{tab:eve2_per_iter}%
\begin{center}
\begin{minipage}{\linewidth}\centering
\perIterCaption{EvE run 2 per-iteration results.}
{\scriptsize
\begin{tabular}{@{}rlrrr@{}}
\toprule
Iter & PE Class & $\overline{e}$ & Best-so-far & $T_{\mathrm{eq}}$ (M) \\
\midrule
0 & \texttt{VanillaICON} & 0.4848 & 0.4848 & 0.0 \\
1 & \texttt{InterpolatedDemoRoleEmbedding} & 0.1208 & \textbf{0.1208} & 2.6 \\
2 & \texttt{StructuredFunctionPE} & 0.1215 & 0.1208 & 4.4 \\
3 & \texttt{StructuredFunctionPE} & 0.1221 & 0.1208 & 5.8 \\
4 & \texttt{StructuredFunctionPE} & 0.1167 & \textbf{0.1167} & 8.4 \\
5 & \texttt{StructuredFunctionPE} & 0.1167 & \textbf{0.1167} & 13.6 \\
6 & \texttt{StructuredFunctionPE} & 0.1167 & \textbf{0.1167} & 16.2 \\
7$^\ast$ & \texttt{StructuredFunctionPE} & 0.1203 & 0.1167 & 18.2 \\
8 & \texttt{StructuredFunctionPE} & 0.1709 & 0.1167 & 22.0 \\
9 & \texttt{StructuredFunctionPE} & 0.1208 & 0.1167 & 23.7 \\
10 & \texttt{StructuredFunctionPE} & 0.1166 & \textbf{0.1166} & 26.0 \\
11 & \texttt{StructuredFunctionPE} & 0.0966 & \textbf{0.0966} & 28.6 \\
12 & \texttt{StructuredFunctionPE} & 0.0966 & \textbf{0.0966} & 30.6 \\
13 & \texttt{StructuredFunctionPE} & 0.0966 & \textbf{0.0966} & 34.2 \\
14 & \texttt{StructuredFunctionPE} & 0.0993 & 0.0966 & 36.0 \\
15 & \texttt{StructuredFunctionPE} & 0.1036 & 0.0966 & 37.5 \\
\bottomrule
\end{tabular}\par
\vspace{2pt}
\raggedright $^\ast$Iteration~7 used one working agent out of two due to a boundary-check violation.\par
}
\end{minipage}
\end{center}

\medskip

\refstepcounter{table}\label{tab:static1_per_iter}%
\begin{center}
\begin{minipage}{\linewidth}\centering
\perIterCaption{Static-Initial run 1 per-iteration results. PE class fixed to \texttt{StructuredDemoRolePE} after the seed iteration.}
{\scriptsize
\begin{tabular}{@{}rlrrr@{}}
\toprule
Iter & PE Class & $\overline{e}$ & Best-so-far & $T_{\mathrm{eq}}$ (M) \\
\midrule
0 & \texttt{VanillaICON} & 0.4848 & 0.4848 & 0.0 \\
1 & \texttt{StructuredDemoRolePE} & 0.2336 & \textbf{0.2336} & 1.7 \\
2 & \texttt{StructuredDemoRolePE} & 0.1357 & \textbf{0.1357} & 3.1 \\
3 & \texttt{StructuredDemoRolePE} & 0.1323 & \textbf{0.1323} & 7.3 \\
4 & \texttt{StructuredDemoRolePE} & 0.1323 & \textbf{0.1323} & 9.0 \\
5 & \texttt{StructuredDemoRolePE} & 0.1466 & 0.1323 & 11.7 \\
6 & \texttt{StructuredDemoRolePE} & 0.1069 & \textbf{0.1069} & 13.4 \\
7 & \texttt{StructuredDemoRolePE} & 0.1215 & 0.1069 & 15.7 \\
8 & \texttt{StructuredDemoRolePE} & 0.1605 & 0.1069 & 18.0 \\
9 & \texttt{StructuredDemoRolePE} & 0.0996 & \textbf{0.0996} & 21.0 \\
10 & \texttt{StructuredDemoRolePE} & 0.0996 & \textbf{0.0996} & 24.8 \\
11 & \texttt{StructuredDemoRolePE} & 0.0996 & \textbf{0.0996} & 27.1 \\
12 & \texttt{StructuredDemoRolePE} & 0.1155 & 0.0996 & 29.9 \\
13 & \texttt{StructuredDemoRolePE} & 0.1018 & 0.0996 & 31.5 \\
14 & \texttt{StructuredDemoRolePE} & 0.1040 & 0.0996 & 33.6 \\
15 & \texttt{StructuredDemoRolePE} & 0.0978 & \textbf{0.0978} & 35.7 \\
\bottomrule
\end{tabular}\par
}
\end{minipage}
\end{center}

\medskip

\refstepcounter{table}\label{tab:static2_per_iter}%
\begin{center}
\begin{minipage}{\linewidth}\centering
\perIterCaption{Static-Initial run 2 per-iteration results. PE class fixed to \texttt{RoleOnlyPE} after the seed iteration.}
{\scriptsize
\begin{tabular}{@{}rlrrr@{}}
\toprule
Iter & PE Class & $\overline{e}$ & Best-so-far & $T_{\mathrm{eq}}$ (M) \\
\midrule
0 & \texttt{VanillaICON} & 0.4848 & 0.4848 & 0.0 \\
1 & \texttt{RoleOnlyPE} & 0.2475 & \textbf{0.2475} & 2.0 \\
2 & \texttt{RoleOnlyPE} & 0.1797 & \textbf{0.1797} & 4.1 \\
3 & \texttt{RoleOnlyPE} & 0.1893 & 0.1797 & 6.0 \\
4 & \texttt{RoleOnlyPE} & 0.1570 & \textbf{0.1570} & 8.9 \\
5 & \texttt{RoleOnlyPE} & 0.1647 & 0.1570 & 10.9 \\
6 & \texttt{RoleOnlyPE} & 0.1865 & 0.1570 & 12.9 \\
7 & \texttt{RoleOnlyPE} & 0.1700 & 0.1570 & 14.9 \\
8 & \texttt{RoleOnlyPE} & 0.1721 & 0.1570 & 19.4 \\
9 & \texttt{RoleOnlyPE} & 0.1746 & 0.1570 & 23.9 \\
10 & \texttt{RoleOnlyPE} & 0.1611 & 0.1570 & 26.2 \\
11 & \texttt{RoleOnlyPE} & 0.1598 & 0.1570 & 28.7 \\
12 & \texttt{RoleOnlyPE} & 0.1770 & 0.1570 & 31.1 \\
13 & \texttt{RoleOnlyPE} & 0.1246 & \textbf{0.1246} & 33.8 \\
14 & \texttt{RoleOnlyPE} & 0.1348 & 0.1246 & 35.4 \\
15 & \texttt{RoleOnlyPE} & 0.1647 & 0.1246 & 37.0 \\
\bottomrule
\end{tabular}\par
}
\end{minipage}
\end{center}

\medskip

\refstepcounter{table}\label{tab:sbo1_per_iter}%
\begin{center}
\begin{minipage}{\linewidth}\centering
\perIterCaption{Static-Final run 1 per-iteration results. PE class fixed to \texttt{StructuredInterpolationPE} after the seed iteration.}
{\scriptsize
\begin{tabular}{@{}rlrrr@{}}
\toprule
Iter & PE Class & $\overline{e}$ & Best-so-far & $T_{\mathrm{eq}}$ (M) \\
\midrule
0 & \texttt{VanillaICON} & 0.4848 & 0.4848 & 0.0 \\
1 & \texttt{StructuredInterpolationPE} & 0.2520 & \textbf{0.2520} & 1.5 \\
2 & \texttt{StructuredInterpolationPE} & 0.1653 & \textbf{0.1653} & 3.9 \\
3 & \texttt{StructuredInterpolationPE} & 0.2695 & 0.1653 & 5.9 \\
4 & \texttt{StructuredInterpolationPE} & 0.2738 & 0.1653 & 9.6 \\
5 & \texttt{StructuredInterpolationPE} & 0.1733 & 0.1653 & 11.1 \\
6 & \texttt{StructuredInterpolationPE} & 0.2058 & 0.1653 & 17.1 \\
7 & \texttt{StructuredInterpolationPE} & 0.1957 & 0.1653 & 19.9 \\
8 & \texttt{StructuredInterpolationPE} & 0.2050 & 0.1653 & 23.2 \\
9 & \texttt{StructuredInterpolationPE} & 0.1989 & 0.1653 & 25.9 \\
10 & \texttt{StructuredInterpolationPE} & 0.1971 & 0.1653 & 29.4 \\
11 & \texttt{StructuredInterpolationPE} & 0.2085 & 0.1653 & 32.0 \\
12 & \texttt{StructuredInterpolationPE} & 0.1678 & 0.1653 & 35.0 \\
13 & \texttt{StructuredInterpolationPE} & 0.1956 & 0.1653 & 37.4 \\
14 & \texttt{StructuredInterpolationPE} & 0.1913 & 0.1653 & 39.5 \\
15 & \texttt{StructuredInterpolationPE} & 0.7127 & 0.1653 & 41.4 \\
\bottomrule
\end{tabular}\par
}
\end{minipage}
\end{center}

\medskip

\refstepcounter{table}\label{tab:sbo2_per_iter}%
\begin{center}
\begin{minipage}{\linewidth}\centering
\perIterCaption{Static-Final run 2 per-iteration results. PE class fixed to \texttt{StructuredDemoRolePE} after the seed iteration.}
{\scriptsize
\begin{tabular}{@{}rlrrr@{}}
\toprule
Iter & PE Class & $\overline{e}$ & Best-so-far & $T_{\mathrm{eq}}$ (M) \\
\midrule
0 & \texttt{VanillaICON} & 0.4848 & 0.4848 & 0.0 \\
1 & \texttt{StructuredDemoRolePE} & 0.1273 & \textbf{0.1273} & 3.3 \\
2 & \texttt{StructuredDemoRolePE} & 0.1628 & 0.1273 & 5.3 \\
3 & \texttt{StructuredDemoRolePE} & 0.1356 & 0.1273 & 9.0 \\
4 & \texttt{StructuredDemoRolePE} & 0.1368 & 0.1273 & 11.1 \\
5 & \texttt{StructuredDemoRolePE} & 0.1356 & 0.1273 & 13.2 \\
6 & \texttt{StructuredDemoRolePE} & 0.1458 & 0.1273 & 15.2 \\
7 & \texttt{StructuredDemoRolePE} & 0.1672 & 0.1273 & 18.1 \\
8 & \texttt{StructuredDemoRolePE} & 0.1217 & \textbf{0.1217} & 19.7 \\
9 & \texttt{StructuredDemoRolePE} & 0.1217 & \textbf{0.1217} & 21.6 \\
10 & \texttt{StructuredDemoRolePE} & 0.1255 & 0.1217 & 23.9 \\
11 & \texttt{StructuredDemoRolePE} & 0.1193 & \textbf{0.1193} & 26.4 \\
12 & \texttt{StructuredDemoRolePE} & 0.1166 & \textbf{0.1166} & 29.1 \\
13 & \texttt{StructuredDemoRolePE} & 0.1193 & 0.1166 & 30.8 \\
14 & \texttt{StructuredDemoRolePE} & 0.1166 & \textbf{0.1166} & 32.5 \\
15 & \texttt{StructuredDemoRolePE} & 0.1422 & 0.1166 & 34.0 \\
\bottomrule
\end{tabular}\par
}
\end{minipage}
\end{center}

\end{document}